# VortexChat: An agentic framework for autonomous multi-objective integrated photonic design

Faqian Chong[1, 2, 3], Yulun Wu[1, 2, 3], Shilong Li[1, 2, 3, 6], Andrew Forbes[4], Hongsheng Chen[1, 2, 3, 5]*, Song Han[1, 2, 3, 6]*

1 Innovative Institute of Electromagnetic Information and Electronic Integration, College of Information Science & Electronic Engineering, Zhejiang University, Hangzhou, 310027, China.

2 ZJU-Hangzhou Global Scientific and Technological Innovation Center, Zhejiang University, Hangzhou 310200, China.

3 International Joint Innovation Center, The Electromagnetics Academy at Zhejiang University, Zhejiang University, Haining 314400, China.

4 School of Physics, University of the Witwatersrand, Private Bag 3, Wits 2050, South Africa.

5 Zhejiang Key Laboratory of Advanced Micro/Nano Electronic Devices & Smart Systems, Jinhua Institute of Zhejiang University, Zhejiang University, Jinhua 321099, China.

6 Zhejiang Key Laboratory of Advanced Micro-nano Transducers Technology, Shaoxing Institute of Zhejiang University, Zhejiang University, Shaoxing 312000, China.

*Corresponding authors: Hongsheng Chen (hansomchen@zju.edu.cn); Song Han (song.han@zju.edu.cn)

## Abstract

The advancement of modern integrated photonics is frequently bottlenecked by device design workflows that rely heavily on manual simulation and expert intuition. While inverse design offers an alternative, it remains constrained by expert supervision and a lack of end-to-end automation. To address these issues, we present VortexChat, an agentic framework for the autonomous, end-to-end inverse design of integrated photonic devices directly from natural language specifications. VortexChat couples a large language model (LLM) decision agent with topology generation, gradient-based refinement, and full-wave electromagnetic simulation. This closed-loop architecture enables the system to iteratively decompose design objectives, orchestrate computational tools, and update strategies based on feedback with minimal human intervention. Constrained by the absolute metrics of the Vortex100 Benchmark, VortexChat autonomously generates devices that strictly meet all predefined performance thresholds without any human-in-the-loop. As an experimental demonstration, we fabricated a broadband terahertz perfect vortex beam multiplexer, autonomously designed by VortexChat, with measurements confirming high-efficiency operation, high mode purity and low inter-channel crosstalk in agreement with full-wave simulations. These results demonstrate that an LLM agent can assume key aspects of expert decision-making in photonic inverse design while maintaining physical fidelity and fabrication feasibility, providing a scalable route towards autonomous design of complex integrated photonic systems.

# Introduction

Integrated photonics[1-3] provides a platform for high-speed communications[4,5] , optical computing[6,7], imaging[8,9] and quantum technologies[10,11] by controlling light propagation[12] and coupling[13] on a chip. Advances in nanofabrication and electromagnetic (EM)-field engineering have enabled photonic crystals[14,15], metamaterials[16,17], metasurfaces[18] and other subwavelength structures[19] to manipulate EM fields at length scales far below the operating wavelength. Yet designing complex integrated photonic structures still relies heavily on human intuition and iterative trial and error. A conventional workflow requires researchers to propose an initial geometry from physical insight, construct the device model, run full-wave simulations and adjust structural parameters according to the simulated response. As illustrated in Fig. 1a, design, modelling, simulation and result analysis each require expert intervention. Inverse design [20,21] offers an important route to this bottleneck. It starts with a prescribed target response and searches the design space for a structure that satisfies the objective, avoiding one-by-one testing of candidate geometries. Existing approaches include heuristic algorithms[22,23], gradient based optimization[24,25] and data-driven deep neural networks[26-29]. Heuristic methods provide broad global search capability but usually require many simulation evaluations[23,30]. Gradient methods can efficiently handle large-scale continuous parameter optimization, but are sensitive to the initial structure, objective function definition and constraint settings[31]. Deep neural networks can learn nonlinear mappings between structures and responses and accelerate exploration of high-dimensional design spaces, but their performance depends strongly on training data quality. The generated structures also often lack clear physical interpretability and manufacturability[32].

Moreover, most inverse design methods remain tool-centered workflows rather than goal-driven autonomous design systems. Researchers must understand EM simulation and optimization algorithms, select appropriate tool chains, configure optimization parameters, monitor convergence and manually adjust the strategy when the algorithm stagnates or the result violates fabrication constraints[25]. Even when the desired optical response is known, designing a practical device structure still requires substantial expert intervention. This dependence limits the accessibility and broader adoption of advanced photonic design. Large language models (LLMs)[33,34] have recently shown strong capabilities in natural language understanding[35], reasoning and planning[36,37] and tool use[38], creating an opportunity to convert complex scientific design tasks into agent-driven automated workflows. In the field of photonics, LLMs have been explored as surrogate models and inverse design engines by using text descriptions of device structures and EM responses[39]. These studies have enabled spectral prediction for relatively simple systems, such as metasurfaces[40], and inverse design of thin-film structures[41]. However, such approaches are often data-intensive and computationally costly, and their performance remains limited for complex free-form photonic structures. A complementary strategy is to use LLMs as orchestration agents that coordinate external solvers, simulation tools and optimization modules with higher physical fidelity. Such frameworks have supported natural language interpretation of design requirements[42], integration of scientific knowledge into design workflows, and coding or simulation assistance for metasurfaces[43,44], optical fibres[42] and laser cavities[45]. Nevertheless, the integration of LLM-based agents with high-complexity photonic optimization and closed-loop physical feedback remains underdeveloped, particularly for multi-objective free-form devices subject to fabrication constraints.

Here we propose VortexChat, an agent-driven automated design framework that implements an end-to-end closed loop from natural language design requirements to multi-objective integrated photonic devices with minimal human intervention (Fig. 1b). Unlike conventional tool-centered workflows requiring constant human monitoring, the core of VortexChat lies in integrating the reasoning and planning of an agent with a specialized tri-tool ecosystem (Fig. 1c). To reduce the sustained expert attention required to operate conventional inverse-design workflows while ensuring fabrication feasibility, the agent dynamically orchestrates OAMDiff (a conditional diffusion model) for rapid, physically plausible topological initialization, Gradient-based Refinement (GBR) for transforming these topologies with continuous permittivity into fabrication-compatible binary structures via adjoint optimization, and StructureTester (ST) for autonomous full-wave simulation feedback. By translating subjective user intents into structured parameters, the agent effectively acts as an expert designer—autonomously interpreting full-wave metrics, coordinating tool execution, closing the feedback loop, and adjusting optimization strategies to guarantee convergence with minimal human intervention.

When operating multi-objective photonic design, we establish the Vortex100 benchmark to evaluate VortexChat and demonstrate a 71% success rate under simultaneous performance constraints of designed devices on efficiency (> 75%), mode purity (> 80%), and bandwidth (> 70% efficiency at each of uniformly sampled wavelengths). Although VortexChat is conceived as a general-purpose photonic synthesis framework, we validate its capability using broadband terahertz perfect-vortex (high-dimensional complex EM field) emitter and its OAM multiplexing via multi-port excitations, where each port feeds into the same inverse-designed region and independently radiates a distinct OAM order, as a representative challenge. By autonomously navigating this design space, VortexChat produces physically valid and fabrication-aware multi-mode THz photonic devices, whose broadband vortex-field generation is further confirmed experimentally. These results establish VortexChat as a versatile framework for autonomous inverse design of complex integrated photonic systems.

# Results

## Agentic integrated photonics design framework

VortexChat is an autonomous optimization framework powered by an open-source LLM[46]. Details of the base model are provided in Methods. The core agent follows the ReAct paradigm[36], in which reasoning and external tool use are interleaved throughout the design process[37]. Unlike a fixed procedural automation workflow, VortexChat continuously integrates the global design objective, current structural parameters, simulation feedback and historical optimization records into its decision-making process. On this basis, the agent dynamically determines tool selection, parameter updates and optimization strategies, enabling the generation of target EM structures under both physical and fabrication constraints.

The automation capability of VortexChat is enabled by a set of carefully designed prompts, provided in the Supplementary Information Sec.1. These prompts define the design objectives, execution order, action boundaries and constraint conditions. Inspired by the Model Context Protocol (MCP) architecture[47], the agent communicates with external tools through structured JSON messages. As shown in Fig. 1c, VortexChat autonomously coordinates three functional modules according to the design target and real-time feedback. OAMDiff is a diffusion model-based topology generation module that rapidly produces physically plausible initial structures, thereby reducing the search space and shortening the subsequent optimization process. GBR is an adjoint gradient based inverse design module that further optimizes the initial topology under prescribed objective functions and fabrication constraints. It transforms a continuous permittivity topology into a fabrication-compatible silicon-air binary structure. ST automatically constructs full-wave EM simulation models and returns quantitative device performance metrics.

During optimization, the agent obtains progress information and intermediate parameters from the Feedback module, and uses this information to adjust the inputs of OAMDiff and GBR. When necessary, it can switch optimization routes or modify the workflow. Because GBR is implemented as a Python-based optimization module, the agent can also intervene at the code level by modifying optimization scripts, objective functions or model parameters to improve efficiency and convergence stability. Through this closed-loop integration of reasoning, generation, simulation and feedback-based refinement, VortexChat realizes autonomous multi-objective EM structure design and outputs fabrication-oriented photonic layouts. More generally, the framework separates decision-making from task-specific physical tools, allowing the same closed-loop design logic to be transferred to other integrated-photonic problems involving different target fields, device geometries, material platforms and performance constraints.

## OAMDiff: Diffusion-based photonics structures generation

The design of photonic structures can be formulated as an inverse problem of establishing a mapping between a target response $O$ and a physical structure $S$, such that the generated structure reproduces the desired EM behavior as accurately as possible. This process can be expressed as[48]:

$$S^{*} = \arg\min_{S \in \mathcal{S}} \mathcal{L}[(F(S), O)] + \mathcal{R}(S) \tag{1}$$

where $\mathcal{S}$ denotes the design space, $F(S)$ represents the EM response produced by structure $S$, $\mathcal{L}[(F(S), O)]$ measures the mismatch between the target and actual responses, and $\mathcal{R}(S)$ denotes regularization terms that enforce physical and fabrication constraints.

As with many design problems of integrated photonics, devices for complex optical field manipulation can be described within the mapping between a target response $O$ and a physical structure $S$. We therefore selected broadband perfect vortex emitter as a representative and experimentally demanding instance of this broader class of inverse design tasks. Perfect vortex beams are high-dimensional complex optical fields with helical phase fronts and annular intensity profiles whose radius is approximately independent of the topological charge[49]. Their field profiles can be expressed as[50]:

$$E_{\ell}^{\mathrm{PV}}(r, \phi) = A_{\ell} \exp\left[-\frac{(r - r_0)^2}{w_0^2}\right] \exp(i\ell\phi) \tag{2}$$

where $(r, \phi)$ are the spatial coordinates and phase, $w_0$ denotes beam waist, and $\ell$ is the azimuthal (topological) charge. The phase term $\phi$ accumulates linearly around the azimuthal direction and integrates to $2\pi\ell$. $A_\ell$ is the amplitude normalization constant. Perfect vortex beam devices are well suited for evaluating automated photonic design frameworks because their design simultaneously involves multiple coupled constraints, including phase matching, annular intensity control, mode purity, conversion efficiency, bandwidth, multi-port balance and inter-channel crosstalk.[49,51] For a multi-port multiplexer, each input port corresponds to a target OAM order, and the device must maintain high efficiency, high mode purity and low crosstalk over a broad spectral range under fabrication constraints. Therefore, we formulate broadband perfect vortex beam multiplexer design as a multi-objective EM structure optimization problem. Definitions of all performance metrics are provided in Methods.

A high-quality initial solution can substantially reduce optimization time and mitigate convergence to poor local optima[31]. To this end, we introduce OAMDiff, a conditional diffusion model that generates continuous EM topologies for integrated photonics devices. OAMDiff serves as the front-end topology generation module of VortexChat. It produces physically plausible initial structures that are already aligned with the target device response. By providing a favorable starting point for subsequent optimization, OAMDiff narrows the search space and reduces inefficient optimization trajectories. More generally, its conditional generation strategy learns a response-to-topology distribution, providing a transferable modelling principle for other photonic design problems when trained with the corresponding response conditions and structural data.

Diffusion models have demonstrated remarkable success in conditional image and video generation tasks, making them well suited for learning the mappings between target device specifications and corresponding topologies. Denoising diffusion probabilistic models (DDPMs) progressively reconstruct a target data distribution from noise[52]. Latent diffusion models (LDMs) further improve computational efficiency and conditional generation stability by performing diffusion in a compressed latent space[53]. This makes them well suited for modelling the highly nonlinear relationship between EM performance and free-form photonic structures. Details of dataset generation and preprocessing for training OAMDiff are provided in Methods.

To reduce the computational burden associated with high-dimensional pixel-space diffusion, we adopt a latent diffusion framework. As illustrated in the upper panel of

Fig. 2a, a VQ-VAE[54] encoder compresses the topology $x$ into a latent representation $z$:

$$z_0 = \text{Encoder}(x) \tag{3}$$

The latent variable $z_0$ is subsequently corrupted through a forward diffusion process that gradually maps the latent distribution to a standard Gaussian distribution $\mathcal{N}(0, I)$, where $I$ is the identity matrix. The architecture and parameters of VQ-VAE can be found in the Supplementary Information Sec.3. A single forward diffusion step is defined as:

$$q(z_t|z_{t-1}) = \mathcal{N}\left(z_t; \sqrt{1-\beta_t} z_{t-1}, \beta_t I\right) \tag{4}$$

Over $T$ diffusion steps, the forward process forms a fixed Markov chain:

$$q(z_T|z_0) = \prod_{t=1}^{T} q(z_t|z_{t-1}) \tag{5}$$

Conditional generation corresponds to reversing this diffusion process through a neural network $p_\theta(z_{t-1}|z_t, c)$ parameterized by $\theta$, where $c$ denotes the conditioning variables. As shown in the lower panel of Fig. 2a, once training is completed, new latent samples are generated through the reverse Markov chain:

$$p_\theta(z_0|c) = \prod_{t=1}^{T} p_\theta(z_{t-1}|z_t, c) \tag{6}$$

$$p_\theta(z_{t-1}|z_t, c) = \mathcal{N}(z_{t-1}; \mu_\theta(z_t, t, c), \delta^2 I) \tag{7}$$

The generated latent topology is subsequently decoded back into pixel space through the VQ-VAE decoder:

$$\tilde{x} = \text{Decoder}(\tilde{z}_0) \tag{8}$$

The variance term $\delta^2$ is given by:

$$\delta^2 = \frac{1 - \overline{\alpha_{t-1}}}{1 - \overline{\alpha_t}} \beta_t I \tag{9}$$

and the mean term is determined by the predicted noise:

$$\mu_\theta(z_t, t, c) = \frac{1}{\sqrt{\alpha_t}} \left( x_t - \frac{1 - \alpha_t}{\sqrt{1 - \overline{\alpha_t}}} \epsilon_\theta(z_t, t, c) \right) \tag{10}$$

The model is trained to predict the Gaussian noise injected at diffusion step $t$. Consequently, the training objective minimizes the mean squared error between the true noise $\epsilon$ and the predicted noise $\epsilon_\theta$:

$$\mathcal{L}(\theta) = E_{z_0, \epsilon \sim \mathbb{N}(0,I), t}[|\epsilon - \epsilon_\theta(z_t, t, c)|^2] \tag{11}$$

where $E$ denotes the mathematical expectation, $\epsilon_\theta$ denotes the noise predicted by the model and $\theta$ denotes the parameters of the neural network.

A symmetric U-shaped convolutional network (U-Net)[55] is used as the conditional noise estimator in OAMDiff. The multi-scale encoder-decoder architecture of U-Net enables the network to capture global structural features while preserving local geometric continuity, making it well suited for the generation of complex topology of the EM structures. As shown in Fig. 2b, OAMDiff adopts a symmetric U-Net to recover continuous topologies. The skip connections in U-Net integrate shallow geometric details with deep semantic features, which is essential for simultaneously satisfying multiple objectives, including vortex beam efficiency, mode purity and operating wavelength. The architecture of denoising U-Net is provided in Methods. As shown in Fig. 2c, cross-attention modules[56] are used to inject target physical conditions into the spatial features of the U-Net, serving as the key pathway for condition-dependent modulation during topology generation. Cross-attention enables fine-grained modulation of spatial locations by the conditional vector[57], allowing the network to directly incorporate multi-objective information, including OAM order, efficiency and wavelength, into the latent representation and thereby improving the physical consistency of the generated topology. Details of the cross-attention are provided in Methods.

To evaluate whether OAMDiff generates usable initial topologies at specified operating wavelengths, we constructed a conditional generation test consistent with the training metrics. The test wavelengths covered 500-1500 μm, and 10 independent topology samples were generated for each wavelength interval. The conditioning vector specified the efficiency and mode purity at the center wavelength for all four output ports. As shown in Figs. 2d and 2e, the initial topologies generated by OAMDiff exhibited stable optical responses across the entire wavelength range. The global average efficiency was approximately 82.6%, and the average efficiency at each wavelength remained above 80.5%, and the efficiency calculation here is based on bidirectional emission due to the symmetric EM condition along the emission directions[58]. The mode purity was higher overall, with an average value of approximately 85.5%, and remained above 84.4% at all wavelengths. Cross wavelength statistics of the four-port average efficiency and purity showed only weak wavelength dependence, indicating that the model maintains stable conditional-generation capability over a broad wavelength range. Comparison of long-term average performance across the four output ports showed small differences in efficiency and purity among ports, substantially below the cross-sample standard deviation of each individual port. This indicates that OAMDiff does not systematically sacrifice any specific output channel. These results show that OAMDiff enables efficient and controllable initial topology generation in a high-dimensional pixel space, with low dispersion in efficiency and purity and strong multi-port balance.

## Physical-based optimization and close-loop configuration

Although OAMDiff can generate continuous topology distributions that satisfy the target optical response, such structures contain spatially varying effective permittivity and therefore cannot be directly fabricated using standard fabrication processes. A complete autonomous design framework should also incorporate physics-based optimization rather than relying only on black box statistical models. To this end, we introduce a fabrication-oriented optimization module, termed Gradient-based Refinement (GBR), into VortexChat. GBR refines the continuous initial topologies generated by OAMDiff into fabrication-compatible binary structures. Details of GBR are provided in the Supplementary Information Sec.5.

As illustrated in Fig. 3a, GBR uses an adjoint gradient optimization algorithm[59] to refine the continuous permittivity distribution generated by OAMDiff. A Heaviside projection filter gradually transforms the smooth continuous permittivity distribution into a silicon-air binary topology. Direct binarization, however, often introduces geometrical artefacts that violate fabrication constraints, including isolated pixels, narrow connections, sharp corners and features below the minimum feature size requirement. To address this problem, GBR incorporates fabrication-aware constraints during the later optimization stages. Penalty terms associated with geometries that are not suitable for fabrication guide the optimization away from fabrication-infeasible regions of the design space. The minimum feature-size constraint is specified by the global design objective provided by VortexChat, allowing the optimization to adapt to different fabrication requirements. Throughout this process, GBR seeks to maximize target optical performance while improving binarization, structural connectivity and fabrication feasibility. Implementation details of binarization and fabrication-aware optimization are provided in the Supplementary Information Sec.6.

The optimization objective of GBR is defined through the mode overlap between the simulated output field and the target perfect vortex mode. For the *i-th* output port, the figure of merit (FOM) is given by:

$$FOM_i = \gamma \frac{|\int \left(E \times H_\ell^{\mathrm{PV}}\right) \cdot dS + \int \left(E_\ell^{\mathrm{PV}} \times H\right) \cdot dS|^2}{\int Re\left(E_\ell^{\mathrm{PV}} \times H_\ell^{\mathrm{PV}}\right) \cdot dS} \tag{12}$$

$E_\ell^{\mathrm{PV}}$ and $H_\ell^{\mathrm{PV}}$ represent the field distributions of the target perfect vortex beam from Eq.1, while $E$ and $H$ are the actual fields obtained from direct simulation, and $\gamma$ is a normalization constant. And the overall FOM is defined as:

$$FOM = \sum_{i=1}^{4} FOM_i \tag{13}$$

As shown in Fig. 3b, the initial continuous topology progressively evolved into a silicon-air binary structure during optimization. The FOM showed several sharp fluctuations along the optimization trajectory, reflecting the dynamic trade-off between optical performance and fabrication constraints. Such fluctuations typically occurred when the Heaviside projection strength increased or when fabrication penalties began to dominate the optimization process. Under these conditions, the optimizer could temporarily sacrifice local optical performance to remove geometrical features that violated fabrication requirements. Despite these transient variations, the optimization ultimately converged to a design that balanced optical performance and manufacturability.

To close the autonomous design loop, VortexChat further incorporates the StructureTester (ST) module. As shown in Fig. 3c, ST is built upon a full-wave EM simulation platform and uses agent-driven scripts to automatically perform device modeling, simulation setup and performance extraction. The automated simulation scripts are provided in the Supplementary Information Sec.6. After execution, ST returns the efficiency and mode purity of the four output ports in a structured format. As illustrated in Fig. 3d, these results are subsequently passed to the Feedback module. Feedback compares the simulated performance with the global design objectives and provides the agent with real-time information regarding optimization progress, including port efficiency, mode purity, convergence behaviour and satisfaction of fabrication constraints. If the design objectives are achieved, VortexChat outputs the final structure together with the corresponding design parameters. Otherwise, the agent autonomously determines the next action based on the returned feedback. Possible actions include adjusting GBR optimization or regenerating a new initial topology through OAMDiff. Through the closed-loop interaction among OAMDiff, GBR, ST and Feedback, VortexChat realizes an autonomous and fabrication-aware design workflow that spans generative topology prediction, physics-based refinement and full-wave validation.

## Vortex100 Benchmark

To systematically and quantitively evaluate VortexChat, we established the Vortex100 Benchmark. Rather than assessing only final optical response, Vortex100 evaluates both device performance and the behavior of the autonomous design workflow. Optical metrics include mode purity, conversion efficiency and operating bandwidth at each output port. Workflow metrics include task-level success rate, the number of tool-calling rounds, wall-clock runtime and total computational cost. Each task is therefore evaluated in terms of center-wavelength performance, bandwidth robustness and computational efficiency. Detailed metric definitions of Vortex100 Benchmark are provided in Methods.

We first evaluated the task-level success rates across the entire benchmark. A task was considered successful only when all three sub-criteria (efficiency, mode purity and bandwidth) were simultaneously satisfied. Under this strict joint criterion, VortexChat achieved an overall success rate of 71% (Fig. 3e). When the three criteria were evaluated separately, the efficiency, purity and bandwidth pass rates were 78%, 72% and 90%, respectively. The substantially higher single-metric pass rates compared with the overall success rate indicate that most failed cases were not caused by a complete breakdown of the design workflow, but by the difficulty of satisfying multiple coupled optical constraints at the same time. In particular, the lower purity pass rate suggests that accurate modal reconstruction is the most restrictive requirement in the benchmark, whereas bandwidth robustness is comparatively easier to maintain once acceptable center-wavelength performance is achieved.

We next compared the performance distributions of successful and failed tasks. For successful tasks, the four-port averaged efficiency and mode purity reached 77.5% and 84.5%, respectively (Fig. 3f, g). For failed tasks, these values decreased to 74.1% and 75.2%. The larger reduction in efficiency, together with stronger efficiency fluctuations in failed cases, indicates that conversion efficiency is the more sensitive indicator separating successful from failed designs. Port-level statistics further showed that the mean efficiencies of the four ports were approximately 76.3%, 76.7%, 76.4% and 77.1%, whereas the corresponding mean purities were approximately 81.6%, 82.5%, 81.2% and 83.1% (Fig. 3h, i).

Tool-calling statistics further revealed the stability of the autonomous workflow. The median number of tool-calling rounds per task was 9, with a range of 5-50 (Fig. 3j, k). Failed tasks required more tool calls on average than successful tasks. This observation suggests that unsuccessful optimization trajectories tend to persist longer before termination, resulting in more tool invocations. Increased tool-calling rounds should therefore be interpreted primarily as a symptom of optimization difficulty rather than as a direct cause of failure.

The median wall-clock runtime was approximately 73 h, with a mean of approximately 82 h and a maximum of approximately 570 h (Fig. 3l). The mean GPU cost was approximately 0.01 GPUh (Fig.3m), and the mean CPU cost was 5.77 CPUh (Fig. 3n). The median total computational cost was approximately $1.16 \times 10^2$ PFLOPs, with a mean of approximately $1.2 \times 10^2$ PFLOPs and a maximum of approximately $7.0 \times 10^2$ PFLOPs (Fig. 3o). The long tail was mainly caused by a small number of tasks in which the agent repeatedly invoked tools or performed low-efficiency iterations after failing to meet the design objectives. Since each task was capped at 50 tool calls, cases in the long-tail regime typically failed to satisfy all design objectives before reaching the termination criterion. A more detailed analysis of the Vortex100 Benchmark is provided in the Supplementary Information Sec.7.

To further assess the contribution of each module, we also performed ablation experiments, as detailed in the Supplementary Information Sec.8. Overall, Vortex100 demonstrates the effectiveness of VortexChat for natural language driven autonomous design of vortex photonic devices. The 71% overall success rate shows that the framework can satisfy efficiency, purity and bandwidth constraints simultaneously in most tasks. The higher pass rates of individual criteria indicate that the main bottleneck lies in joint multi-objective optimization rather than failure of a single metric. The port-level results further demonstrate that VortexChat maintains good multi-port balance, with failed tasks primarily limited by insufficient global performance margin. Notably, Vortex100 is currently used to evaluate autonomous photonic design. Because it jointly assesses objective satisfaction, computational efficiency and agent behaviour under multiple design constraints, the evaluation framework can therefore be extended to other domains involving autonomous, iterative and tool-assisted design.

## End-to-end autonomous design of a broadband terahertz vortex beam multiplexer

To validate the end-to-end autonomous design capability, a human researcher specified the objective solely through natural language, as shown in Fig. 4a. VortexChat first parsed this natural language request into structured global design parameters, and then automatically initiated the subsequent design workflow. The device layout is shown by Fig. 4b. The design region has a lateral size of 5000 μm × 5000 μm and a thickness of 200 μm, and consists of a silicon-air binary material. Four waveguide ports serve as independent excitation channels, each feeding into the same inverse-designed region. Excitation from each port independently produces a distinct OAM mode from the shared radiating aperture. Specifically, excitation from the top, right, bottom and left ports generate target OAM modes of $\ell = 1$, $\ell = 2$, $\ell = 3$ and $\ell = 4$, respectively. This task requires the structure to maintain stable responses across multiple ports, multiple topological charges and a broad wavelength range, thereby providing a stringent test of autonomous multi-objective design capability. During the design process, VortexChat first invoked OAMDiff to generate an initial topology with a continuous permittivity distribution according to the global design objective. This topology was then passed to the GBR module for manufacturability-aware gradient refinement, progressively transforming the structure into a fabricable silicon-air binary layout. After the first round of GBR optimization, ST automatically constructed the full-wave simulation model and returned the efficiency, mode purity and bandwidth metrics for each port. Because the initial refined structure did not fully satisfy the target constraints, the agent adjusted the GBR optimization parameters according to the feedback and launched a second refinement round. After the second optimization, the performance metrics returned by ST met the predefined design requirements. VortexChat therefore terminated the iteration and output the final device structure together with its performance parameters. This process proceeded without manual intervention during the optimization loop, demonstrating a closed-loop autonomous design capability that spans natural language requirement parsing, generative topology prediction, physics-constrained optimization and full-wave simulation validation.

Full-wave simulations verified the broadband multiplexing performance of the autonomously generated device. As shown in Fig. 4c, within the target 750 – 900 μm band, all four ports produced clear annular intensity profiles and continuous helical phase fronts, confirming stable emission of the corresponding vortex beam orders. The phase winding of each mode was consistent with its target topological charge, indicating that the structure not only achieved efficient energy coupling but also correctly reconstructed the phase characteristics of the desired OAM modes. Simulated efficiency analysis further showed that the device maintained a relatively flat broadband response (Fig. 4d). As shown in Fig. 4e, the simulated mode purity of all four target OAM modes remained above 85% across the operating band. Further simulated modal isolation analysis showed that the maximum crosstalk between different OAM channels was ~5.0% (Fig. 4f), confirming that the multiplexer maintains good channel isolation during parallel operation of multiple vortex beam orders.

The inverse-designed terahertz photonic devices were fabricated using a standard photolithographic masking process. The devices were patterned on single crystal silicon wafers with a thickness of 200 μm. Details of the fabrication process are provided in the Supplementary Information Sec.10. Fig. 5a shows an optical microscope image of a representative device, consisting of a tapered waveguide coupler, an effective medium waveguide with a width of 210 μm, and the inverse-designed region. The effective medium theory and waveguide parameters are described in the Supplementary Information Sec.9. The fabricated devices were characterized using a custom built 2D field-scanning system integrated with vector network analyzer extension modules (VNAX). Details of the measurement platform and experimental procedure are provided in the Supplementary Information Sec.11. Fig. 5b shows the experimentally measured field-amplitude and phase distributions. The measured fields agree with the numerical simulations, confirming that the inverse-designed terahertz photonic devices can robustly emit high-quality vortex beams and reconstruct the annular intensity profiles and helical phase fronts of the target modes. Fig. 5c presents the experimentally measured unidirectional emission efficiencies of devices with different OAM orders across the operating bandwidth. Because the devices radiate bidirectionally, the theoretical maximum efficiency for unidirectional emission is 50% (−3 dB); all efficiencies reported here are therefore referenced to the input power. The experimental efficiencies remain within the range of −5.0 to −4.6 dB across the operating band. Experimentally, the best efficiency of the +4-order device reaches −4.6 dB at a wavelength of 900 μm. As shown in Fig. 5d, the measured mode purities of the four target OAM modes remain above 84% throughout the operating band, indicating strong agreement between the measured output fields and the target perfect vortex modes. Further measured modal isolation analysis shows that the maximum crosstalk between different OAM channels is 5.2% (Fig. 5e), confirming that the multiplexer maintains good channel isolation during parallel operation of multiple vortex beam orders.

## Conclusions and discussion

We developed VortexChat, an autonomous agentic framework for the end-to-end design of integrated photonic devices. By coordinating three specialized tools, VortexChat autonomously translates natural-language specifications into executable multi-objective design workflows and fabrication-compatible device structures. The framework has been systematically evaluated on complex optical-field manipulation tasks and has successfully designed broadband terahertz perfect vortex beam multi-port multiplexers under simultaneous constraints on efficiency, mode purity, and bandwidth. Fabrication and experimental characterization have further confirmed that the generated structures can be realized using practical processes and can reproduce the intended optical-field responses. These results demonstrate that VortexChat reduces the dependence of complex integrated-photonic design on continuous expert supervision while maintaining physical fidelity and fabrication feasibility. More broadly, the framework has established a practical route for applying LLM agent-driven tool orchestration to multi-objective device design and provides a scalable foundation for the autonomous development of complex integrated photonic systems.

From a user and application perspective, VortexChat can be integrated with multimodal large language models through APIs and deployed on cloud servers, allowing non-expert users to access high-performance computing resources through natural language interaction and complete the design of complex integrated photonic devices. This mode of operation may lower the barrier to professional photonic device design and open new commercialization pathways for cloud-based computer-aided design, rapid prototyping and on-demand device generation. From the perspective of framework development, equipping LLM-driven agents with more efficient, specialized and physically reliable tools will be essential for improving the efficiency of autonomous design systems and expanding their capability boundaries. For example, combining specialized agents with fast and physically accurate surrogate solvers could substantially accelerate applied physics research in both industry and academia. At the same time, foundation models with stronger reasoning capabilities will further improve automated design frameworks in task decomposition, constraint interpretation, objective trade-off and design correction. More broadly, LLM-based agents can not only generate new scientific hypotheses, but also revise these hypotheses in real time within closed loops involving simulation, optimization and experimental feedback. As agents, physical solvers, experimental automation and manufacturing platforms become increasingly integrated, future autonomous design systems may continuously evolve and discover new physical structures and functional devices beyond human intuition. These developments suggest a promising direction for scientific research and engineering design, while further validation across broader device classes and experimental settings will be necessary.

# Methods

## Base model used in VortexChat

The base model used in VortexChat was LLaMA 3.1-70B. During inference, the temperature was set to 0.2, and the context length was set to 64k tokens.

## Collection and preprocessing of OAMDiff training samples

Each training sample was represented by a continuous permittivity matrix containing 121 × 121 elements, where 0 denotes air and 1 denotes silicon. Each topology was generated through Staged Annealing Topology Optimization (SATO)[58] iterative optimization, with a device thickness of 200 μm, a width of 5000 μm and a length of 5000 μm. These topologies were cross-validated using the commercial full-wave simulation software Ansys Lumerical FDTD. Perfectly matched layer boundary conditions and mode port excitation were used, and a field monitor was placed at a distance of three wavelengths above the sample surface to evaluate the output field. The efficiency and mode purity extracted from the simulations were used as labels for the corresponding topology, forming the training dataset. The conditioning variables consist of the efficiency, mode purity and operating wavelength range of the target perfect vortex beam. Specifically, the conditioning vector is defined by the efficiencies and mode purities of the four output ports at the center wavelength. Details of the data collection and labeling workflow are provided in the Supplementary Information Sec.2.

## The architecture of U-net for denoising in OAMDiff

In OAMDiff, multi-scale feature fusion is achieved through channel expansion and contraction, cross-layer skip connections, timestep encoding and conditional embedding. Moreover, the U-Net takes the noisy latent variable $z_t$, the diffusion timestep $t$ and the target-performance condition $c$ as inputs, and outputs the estimated noise $\epsilon_\theta(z_t, t, c)$. During training, the model learns to identify and remove noise from the latent representation under different noise levels and target conditions. The target performance vector $c$ is first mapped to a fixed-dimensional conditional embedding through a multilayer perceptron, and is then jointly injected into the network backbone together with the sinusoidal embedding of the diffusion timestep $t$. The input latent feature $\mathrm{z_t} \in \mathrm{R}^{\mathrm{B\times E\times S\times S}}$ is first mapped by a convolutional layer to the base channel width $\mathrm{C_b}$, and then passed through a four-stage downsampling path. In each UNetBlock, several TimedResBlocks[60] are stacked. The detailed structure of TimedResBlock is provided in the Supplementary Information Sec 4. After group normalization and nonlinear activation, two convolutional layers are applied. The timestep embedding is projected to the channel dimension through a linear layer and added pixel-wise to the feature map, allowing the network to distinguish the noise levels associated with different diffusion timesteps. This output is further combined with the shortcut residual connection, forming a pre-normalized residual unit. The up-sampling path symmetrically restores the feature maps to the original base channel width $C_b$, while cross-layer skip connections preserve shallow structural details. Finally, a convolutional layer maps the features back to the latent channel dimension $E$, producing the conditional noise estimate $\epsilon_\theta(z_t, t, c)$.

**Cross-attention in OAMDiff**

For a network block with attention enabled, we denote the feature output from the preceding layer as $h \in R^{B\times C\times H\times W}$. This feature is flattened into a spatial sequence $X \in R^{B\times N\times C}$, where $N = H \times W$. The spatial feature is linearly projected to form the query matrix $Q$, which represents how different spatial positions query the conditional information Meanwhile, the target-performance condition vector $c$ is embedded into $C_{Ctx} \in R^{B\times N\times D_e}$, and is further linearly projected to form the key and value matrices:

$$Q = XW_Q, K = C_{ctx}W_K, V = C_{ctx}W_V \quad (14)$$

Here, $W_Q \in R^{C\times D_i}$ and $W_K, W_V \in R^{D_e\times D_i}$ are learnable projection matrices. The internal dimension is defined as $D_i = Gd_h$, where the number of attention heads is $G = 8$ and the dimension of each head is $d_h = 64$. The correlation between spatial features and conditional embeddings is then computed through scaled dot-product attention[56]:

$$\text{Attn}(Q, K, V) = \text{softmax}\left(\frac{QK^\top}{d_h}\right)V \quad (15)$$

This operation allows each spatial position to select the most relevant physical condition information according to its current latent feature state and to feed this condition dependent modulation signal back into the topology generation process. The outputs from multiple heads are concatenated, linearly projected and reshaped to the same spatial dimensions as $h$. The attention-modulated feature is then added to the original feature through a residual connection. The fused feature subsequently passes through group normalization and a following TimedResBlock to further extract local geometric features. In this way, the cross-attention module explicitly injects global design objectives into the spatial feature representation without disrupting the local inductive bias of the convolutional backbone.

In OAMDiff, cross-attention is mainly deployed in the intermediate and low-resolution stages of the U-Net encoder and in the corresponding decoder stages. At these levels, the feature maps have larger receptive fields and higher channel capacity, making them suitable for modeling long range dependencies between topology and target electromagnetic responses. By contrast, the highest resolution layers rely primarily on convolutional operations and timestep embeddings, which reduces computational cost while preserving boundary continuity, local connectivity and fabrication relevant geometric details. Details of model efficiency and training resource consumption are provided in the Supplementary Information Sec.4.

**Vortex100 Benchmark**

The Vortex100 Benchmark comprises 100 natural language driven vortex device design tasks. The target operating range from 500-1500 μm was divided into ten 100 μm wide wavelength bands. Each task consists of a user prompt describing the target operating band, OAM orders, port excitation configuration, efficiency threshold, mode purity threshold and device size constraint. To emulate variations in user expression, each task uses a different natural language prompt while preserving the same underlying structured constraints. During evaluation, the system receives only the natural language user prompt and must output the corresponding design parameters. All tasks share the same design constraints except for the target operating band. The device is required to generate four vortex modes in parallel, with target OAM orders of $\ell = [1,2,3,4]$, and all four input ports are excited. Efficiency success is defined as all four ports having center wavelength efficiencies greater than 75%. Purity success is defined as all four ports having center wavelength mode purities greater than 80%. Bandwidth success is defined as all 11 uniformly sampled wavelength points within the target band having port efficiencies greater than 70%. A task is considered successful only when all three criteria are satisfied. The minimum feature size was constrained to 20 μm. The reported benchmark metrics include task success rate, port efficiency, mode purity, bandwidth robustness, number of tool-calling rounds, runtime and computational resource consumption. Computational cost was recorded on a platform equipped with a single NVIDIA GeForce RTX 4090 GPU and a single Intel Xeon Platinum 8488C CPU with 48 cores. For each task, we recorded GPU hours, CPU hours, four-task-parallel wall-clock runtime and estimated FLOPs. The wall-clock runtime represents the actual elapsed time required to complete an individual task when four tasks were executed in parallel. GPU hours and CPU hours represent the accumulated GPU and CPU execution times, respectively. FLOPs were estimated by multiplying the hardware execution time by its theoretical vectorized peak floating-point performance.

**Definitions of efficiency and mode purity**

During training-sample collection, preprocessing and generated structure evaluation in the StructureTester module, full-wave simulations were performed using the commercial full-wave simulation software Ansys Lumerical FDTD. The electric field distributions were extracted from the monitor planes and used to calculate the device performance metrics.

The total power flux through the monitor plane is defined as:

$$P_{in/out} = \iint \boldsymbol{S} \cdot \hat{n}\, dS \tag{16}$$

where $\boldsymbol{S}$ is the time averaged Poynting vector, $\hat{n}$ is the normal vector of the surface element, and $S$ is the area of the output monitor plane. By normalizing the output power with the total input power in the waveguide coupling region, the simulation efficiency is defined as:

$$\eta_{\text{sim}} = \frac{P_{out}}{P_{in}} \tag{17}$$

To evaluate mode purity, we used a Fourier transform method to analyze the angular field distribution of the emitted beam. The electric field on the observation plane at a sampling radius is denoted as $E(\phi)$, where $\phi$ is the azimuthal coordinate. The complex amplitude coefficient of the OAM component with order $\ell$ is obtained as:

$$A_\ell = \frac{1}{2\pi} \int_{-\pi}^{\pi} E(\phi) \cdot e^{-i\ell\phi}\, d\phi \tag{18}$$

where $A_\ell$ is the complex amplitude coefficient corresponding to each OAM mode $\ell$. Further, for a specific OAM mode of order $\ell_0$, the mode purity is defined as the ratio of the energy of that mode to the total energy of all modes:

$$Purity(\ell_0) = \frac{\left|A_{\ell_0}\right|^2}{\Sigma_{\ell=\ell_a}^{\ell_b} |A_\ell|^2} \tag{19}$$

where $[\ell_a, \ell_b]$ represents the range of OAM mode orders considered in the analysis. In this work, the range is typically taken as $\ell \in [-10,10]$.

**Code & Data Availability**

The code for VortexChat and the calculated and experimental data supporting the findings of this study will be made publicly available on GitHub upon publication of the article.

## Acknowledgements

We acknowledge the National Key R&D Program of China under Grant 2024YFB2808200, the National Natural Science Foundation of China under Grant 62475230, the Zhejiang Provincial Leading Innovation and Entrepreneurship Team

Program—Young Scientist Innovation Team under Grant 2025R01003, and the Excellent Young Scientists Fund Program (Overseas) of China.

## Author contributions

S.H., and F.Q.C. conceived the project and designed the experiments. F.Q.C. developed VortexChat architecture, wrote the code, implemented the AI agents, developed the Benchmark data analysis, and performed experimental measurements. F.Q.C., Y.L.W. and S.H. analyzed the experimental data. F.Q.C., S.L.L., A.F., H.S.C., and S.H. discussed the results, provided critical feedback and edited the manuscript. S.H. acquired financial support for this project. S. H., and H.S.C. jointly supervised the project.

## Conflict of Interest

The authors declare no conflicts of interest.

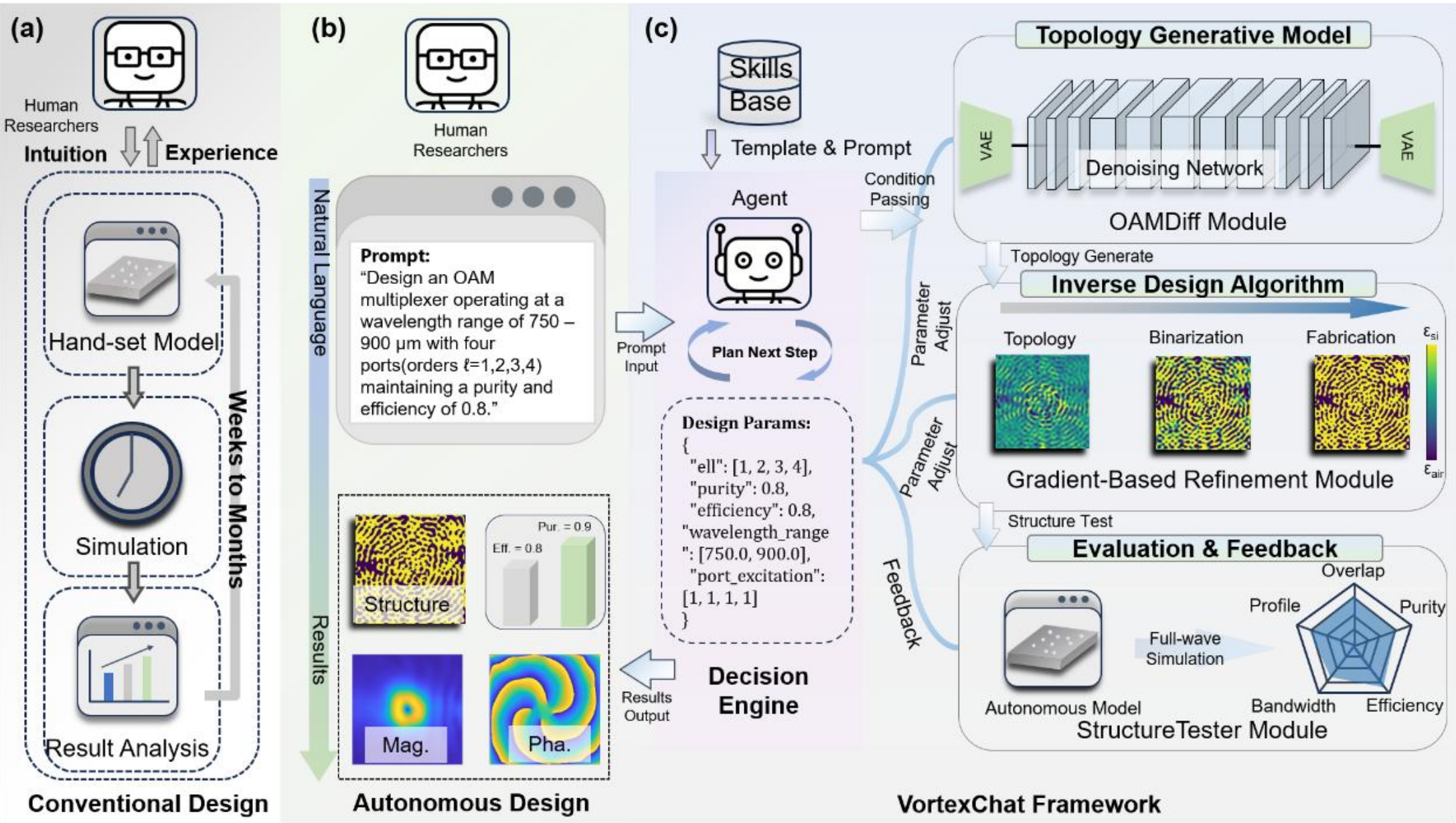

**Fig. 1 | Autonomous design agent framework of VortexChat.**

**a,** Conventional design workflow for integrated photonics. Researchers manually construct models, define structural parameters, and iteratively optimize the design through numerical simulation and result analysis depending on their intuition and experience.

**b,** Natural language driven autonomous design workflow. After receiving the target design requirements as a natural language prompt from human researchers, VortexChat automatically generates the target design parameters and evaluates key performance metrics, including mode purity, energy conversion efficiency and target response.

**c,** Autonomous design agent architecture of VortexChat. The framework consists of a decision engine based on a base model and three core tools: an initial topology generative model, an inverse design refinement algorithm and a performance evaluation feedback module. The agent converts the natural language prompt into structured global optimization objectives, selects tools according to the current design state, adjusts tool-calling parameters and executes a closed loop of generation, evaluation and optimization.

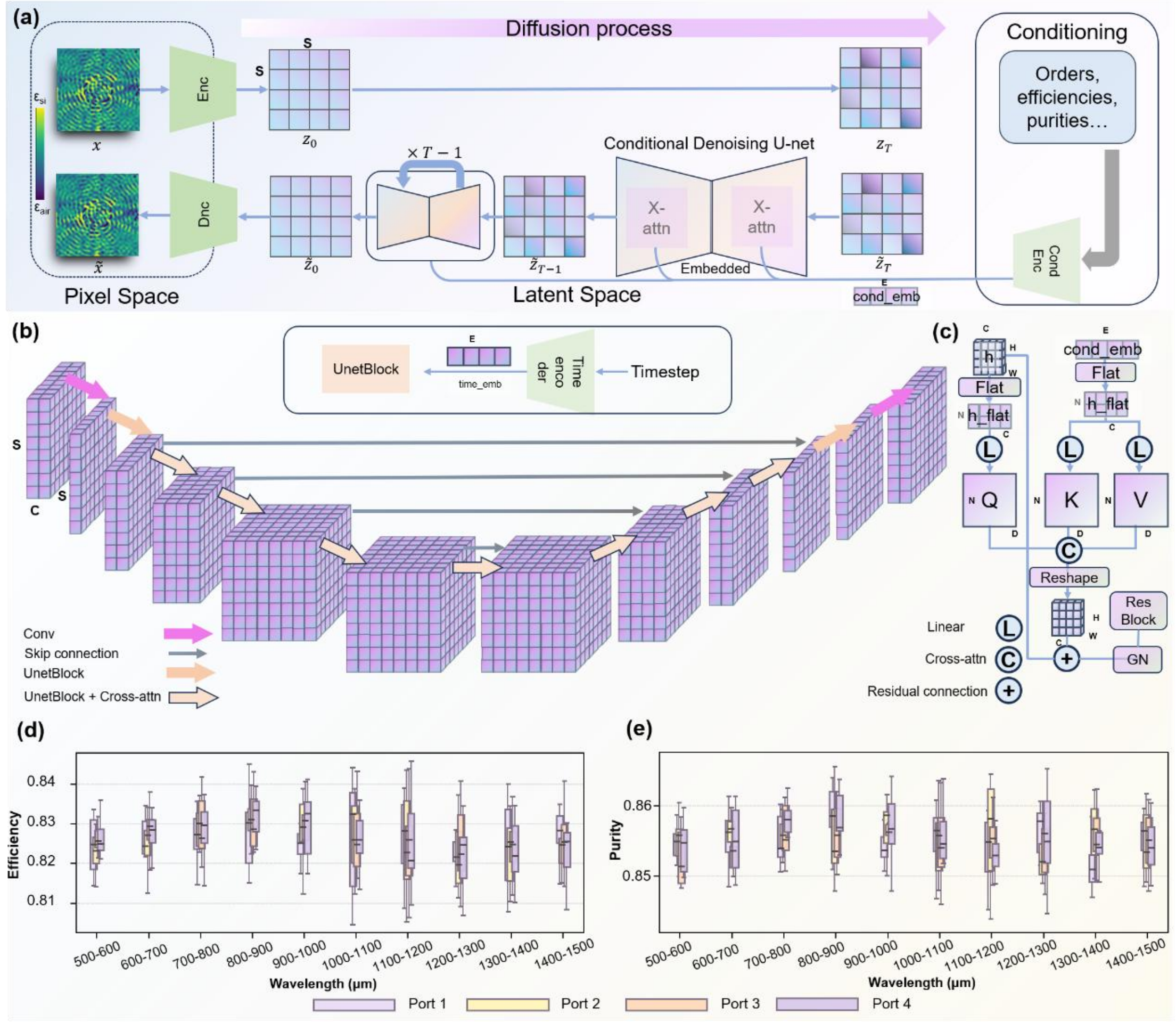

**Fig. 2 | Conditional latent diffusion framework and validation workflow of OAMDiff.**

**a,** Conditional latent diffusion generation process of OAMDiff. The OAMDiff consists of a VQ-VAE as encoder and decoder, and a latent diffusion model. During training, the input topology $x$ is encoded into a latent variable $z_0$ by pretrained VQ-VAE, followed by a forward diffusion process that progressively adds Gaussian noise. The conditional denoising network predicts the diffusion target from the noisy latent variable, diffusion timestep and condition embedding. The condition vector contains the OAM order, mode purity, energy conversion efficiency, operating wavelength and port excitation parameters. During inference, the model samples an initial latent variable from a standard Gaussian distribution latent variable $\tilde{z}_T$, performs reverse denoising under the given conditions and decodes the resulting latent variable $\tilde{z}_0$ into the target topology $\tilde{x}$ using the VQ-VAE decoder.

**b,** Conditional U-Net network for latent space denoising. The noisy latent variable is first mapped to the base channel dimension by an input convolution, followed by four down-sampling stages, a middle stage and four up-sampling stages with skip connections between corresponding levels. Each U-Net block contains time-conditioned residual blocks (TimedResBlock), in which the diffusion timestep is encoded using sinusoidal positional encoding and injected into the block through a multilayer perceptron. Cross-attention modules are placed in the middle stage and deeper U-Net blocks to introduce the condition embedding. Magenta arrows denote convolutional layers, grey arrows denote skip connections and orange arrows denote U-Net blocks; bordered U-Net blocks include cross-attention modules.

**c,** Cross-attention mechanism for condition injection. The topology feature map is flattened into a sequence and used as the query. The condition embedding is spatially replicated and linearly projected to generate the key and value. The attention output is linearly projected, reshaped to the original spatial dimensions and added back to the topology feature map through a residual connection.

**d, e,** Full-wave simulation validation of topologies generated by OAMDiff. The generated continuous permittivity topologies are simulated over the operating wavelength ranges, and the energy conversion efficiency (d) and mode purity (e) are evaluated at the center wavelength.

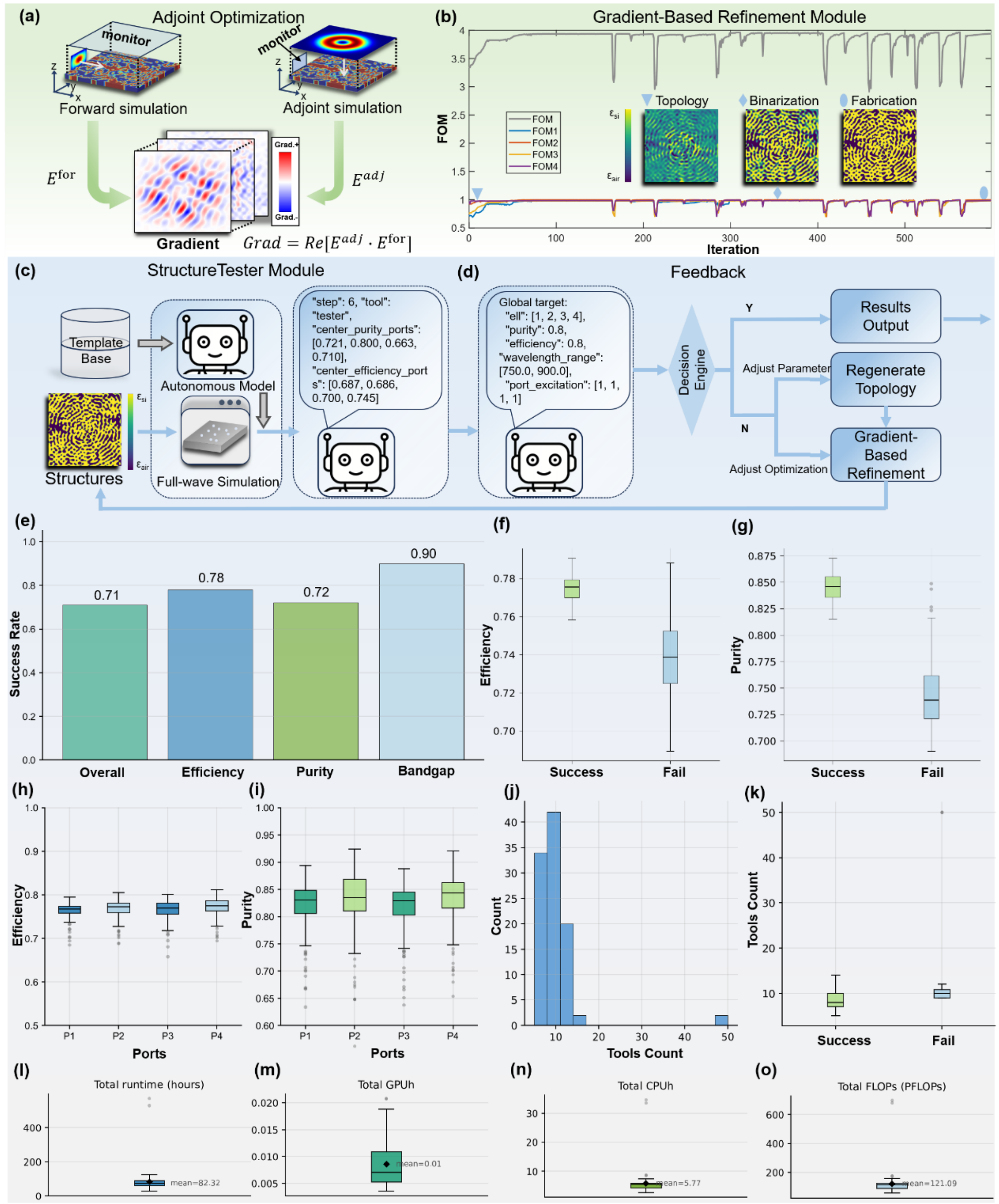

**Fig. 3 | Gradient-based refinement (GBR) module, StructureTester (ST) module and Vortex100 benchmark evaluation in VortexChat.**

**a,** Adjoint gradient calculation in the GBR module. The forward simulation is excited from the input port to obtain the forward electric field, whereas the adjoint simulation injects the complex conjugated field associated with the target figure of merit (FOM) from the output port.

**b,** Optimization workflow of the GBR module. GBR refines the initial topology generated by OAMDiff using the FOM as the optimization objective. Triangles denote the initial topology and generation passed to GBR, diamonds denote the structure and iteration number after binarization, and ellipses denote the structure and iteration number after fabrication-aware optimization.

**c,** Automated modeling and simulation workflow of the ST module. ST receives the structure under test and uses an automated modeling template base to configure modeling parameters, generate simulation files, run full-wave simulations and extract key performance metrics.

**d,** Feedback module and closed-loop decision process. The feedback module compares the performance obtained from ST with the global optimization objectives. If the structure satisfies the target constraints, the system outputs the final design. Otherwise, the decision engine adjusts the condition parameters and calls OAMDiff to regenerate a topology or adjusts GBR optimization for further optimization.

**e,** Overall evaluation results of the Vortex100 benchmark. A total of 100 design tasks is evaluated in terms of overall success rate, efficiency success rate, purity success rate and bandwidth success rate.

**f, g,** Energy conversion efficiency (f) and mode purity (g) of the final output structures at the center wavelength. Results are grouped into successful and failed tasks.

**h, i,** Energy conversion efficiency (h) and mode purity (i) at the center wavelength for different output ports.

**j, k,** Tool-calling statistics for the 100 tasks. The distribution of tool-calling rounds (j) across all tasks; The number of tool-calling rounds (k) between successful and failed tasks.

**l-o,** Computational resource consumption of the autonomous design tasks, including total runtime (l), GPU hours (m), CPU hours (n) and floating-point operations (FLOPs) (o).

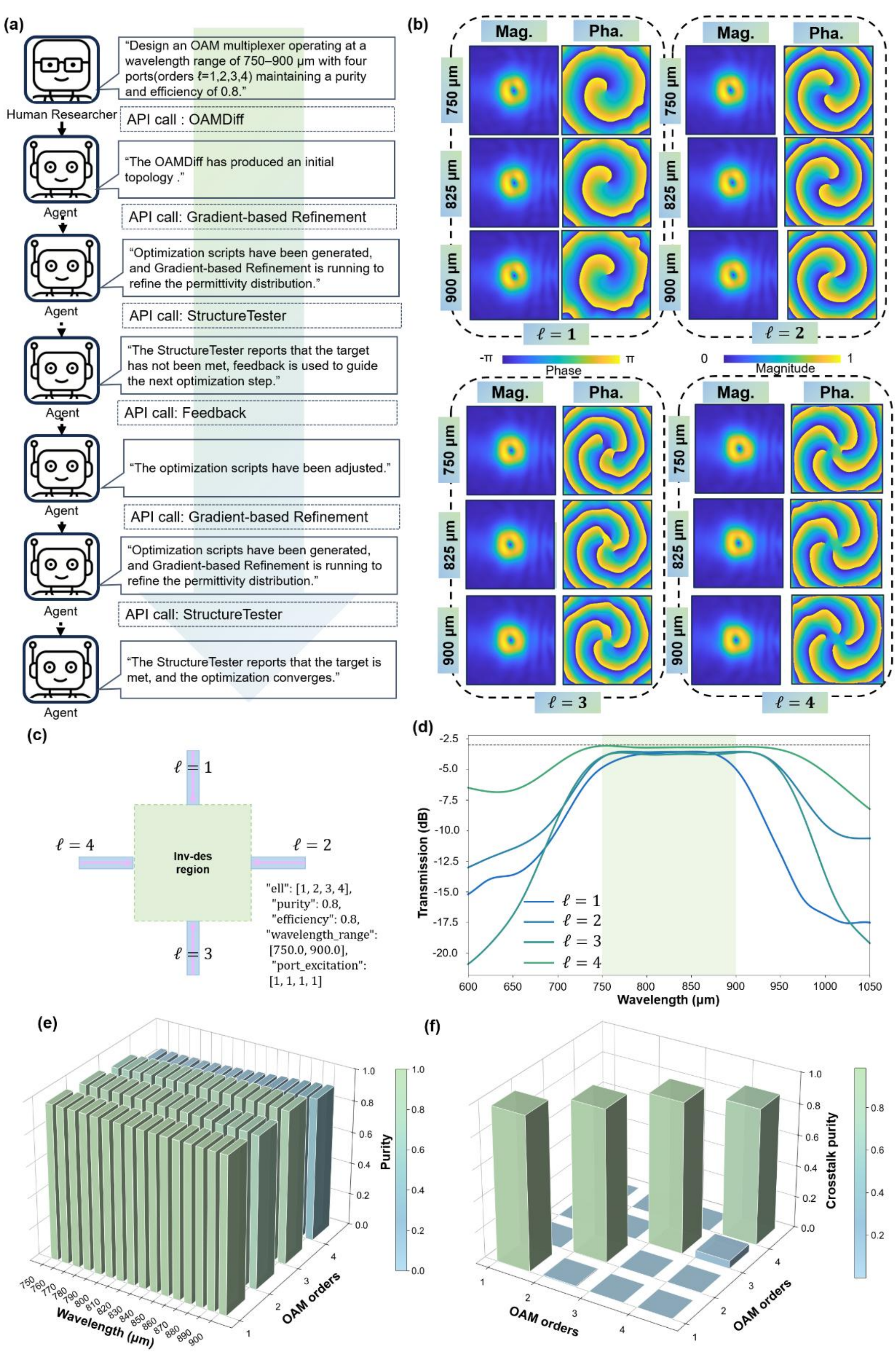

**Fig. 4 | Automated design of a terahertz perfect vortex beam multiplexer using VortexChat.**

**a,** Human-agent interaction and tool calls during the autonomous design process. A researcher provides the design requirements for a terahertz vortex beam multiplexer in natural language. VortexChat parses the design constraints and autonomously calls tools during the design process.

**b,** Field distributions of vortex beams emitted by the designed multiplexer at different operating wavelengths, including amplitude and phase distributions.

**c,** Device concept and structured design objectives of the terahertz vortex beam multiplexer. The design objectives include the operating band, target OAM modes, port configuration, mode purity, energy conversion efficiency and bandwidth.

**d,** Broadband response of the designed multiplexer within the target wavelength range. The green region denotes the predefined target band.

**e,** Mode crosstalk between different OAM orders in the multiplexer.

**f,** OAM mode purity of different topological-charge channels over different wavelength ranges.

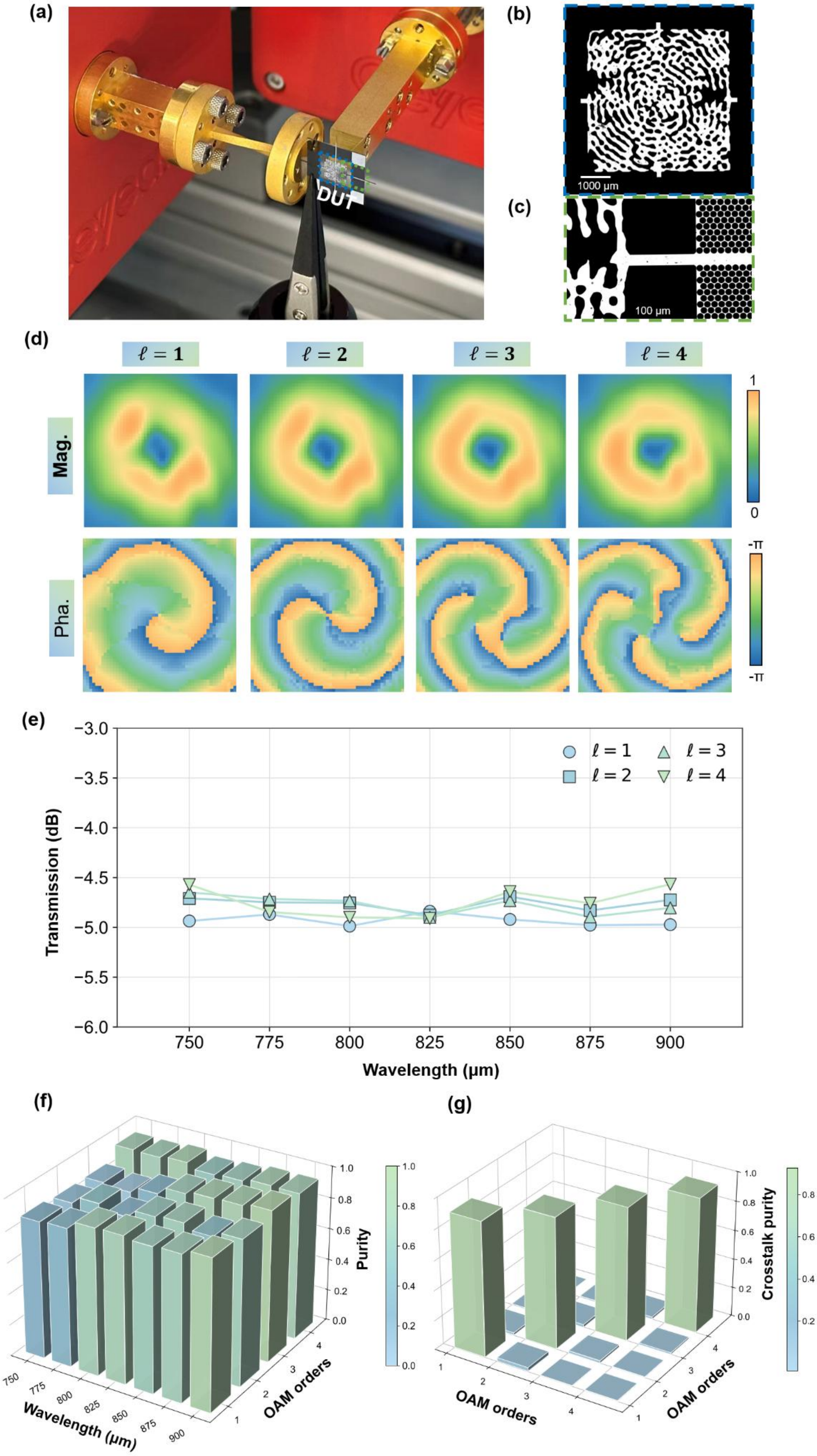

(a)
DUT
(b)
1000 μm
(c)
100 μm
(d)
$\ell = 1$
$\ell = 2$
$\ell = 3$
$\ell = 4$
Mag.
Pha.
1
0
-π
-π
(e)
Transmission (dB)
Wavelength (μm)
$\ell = 1$
$\ell = 2$
$\ell = 3$
$\ell = 4$
(f)
Purity
Wavelength (μm)
OAM orders
(g)
Crosstalk purity
OAM orders
OAM orders

**Fig. 5 | Measured results of broadband integrated terahertz perfect vortex beam multiplexer designed by VortexChat.**

**a,** Experimental terahertz field-scanning measurement setup. The area enclosed by the blue dashed box corresponds to the design area, and green one corresponds to the effective medium waveguide support structures.

**b,** Microscope image of the designed region.

**c,** Microscope image of the effective medium waveguide and subwavelength support structures.

**d,** Measured magnitude and phase distributions of perfect vortex beams with different topological charges at the center wavelength.

**e,** Experimentally measured energy conversion efficiency.

**f,** Experimentally measured OAM mode purity over different wavelength range.

**g,** Experimentally measured mode crosstalk between different OAM channels.